\documentclass[conference]{IEEEtran}
\IEEEoverridecommandlockouts

\usepackage{cite}
\usepackage{amsmath,amssymb,amsfonts}
\usepackage{algorithmic}
\usepackage{graphicx}
\usepackage{textcomp}
\usepackage{xcolor}

\usepackage{xcolor}
\usepackage{listings}
\usepackage{subfigure}
\usepackage{amsfonts}
\usepackage{url}
\usepackage{blindtext}
\usepackage{verbatim}
\usepackage{dirtytalk}
\usepackage{hyperref}

\def\BibTeX{{\rm B\kern-.05em{\sc i\kern-.025em b}\kern-.08em
    T\kern-.1667em\lower.7ex\hbox{E}\kern-.125emX}}
\begin{document}

\title{Benchmarking End-to-End Behavioural Cloning on Video Games}

\author{
    \IEEEauthorblockN{Anssi Kanervisto*}
    \IEEEauthorblockA{\textit{School of Computing} \\
    \textit{University of Eastern Finland}\\
    Joensuu, Finland \\
    anssk@uef.fi}
\and
    \IEEEauthorblockN{Joonas Pussinen*}
    \IEEEauthorblockA{\textit{School of Computing} \\
    \textit{University of Eastern Finland}\\
    Joensuu, Finland \\
    joopu@student.uef.fi}
\and
    \IEEEauthorblockN{Ville Hautam\"aki}
    \IEEEauthorblockA{\textit{School of Computing} \\
    \textit{University of Eastern Finland}\\
    Joensuu, Finland \\
    villeh@uef.fi}

    \thanks{*Equal contribution, alphabetical ordering. This research was partially funded by the Academy of Finland (grant \#313970). We gratefully acknowledge the support of NVIDIA Corporation with the donation of the Titan Xp GPU used for this research. We thank the reviewers for extensive comments used to improve the final version of this paper.
    
    \textcopyright 2020 IEEE.  Personal use of this material is permitted.  Permission from IEEE must be obtained for all other uses, in any current or future media, including reprinting/republishing this material for advertising or promotional purposes, creating new collective works, for resale or redistribution to servers or lists, or reuse of any copyrighted component of this work in other works.}
}


\maketitle

\begin{abstract}
Behavioural cloning, where a computer is taught to perform a task based on demonstrations, has been successfully applied to various video games and robotics tasks, with and without reinforcement learning. This also includes end-to-end approaches, where a computer plays a video game like humans do: by looking at the image displayed on the screen, and sending keystrokes to the game. As a general approach to playing video games, this has many inviting properties: no need for specialized modifications to the game, no lengthy training sessions and the ability to re-use the same tools across different games. However, related work includes game-specific engineering to achieve the results. We take a step towards a general approach and study the general applicability of behavioural cloning on twelve video games, including six modern video games (published after 2010), by using human demonstrations as training data. Our results show that these agents cannot match humans in raw performance but do learn basic dynamics and rules. We also demonstrate how the quality of the data matters, and how recording data from humans is subject to a state-action mismatch, due to human reflexes.
\end{abstract}


\section{Introduction}
\label{sec:intro}
    
    

    \begin{figure*}[ht]
    \includegraphics[width=\linewidth]{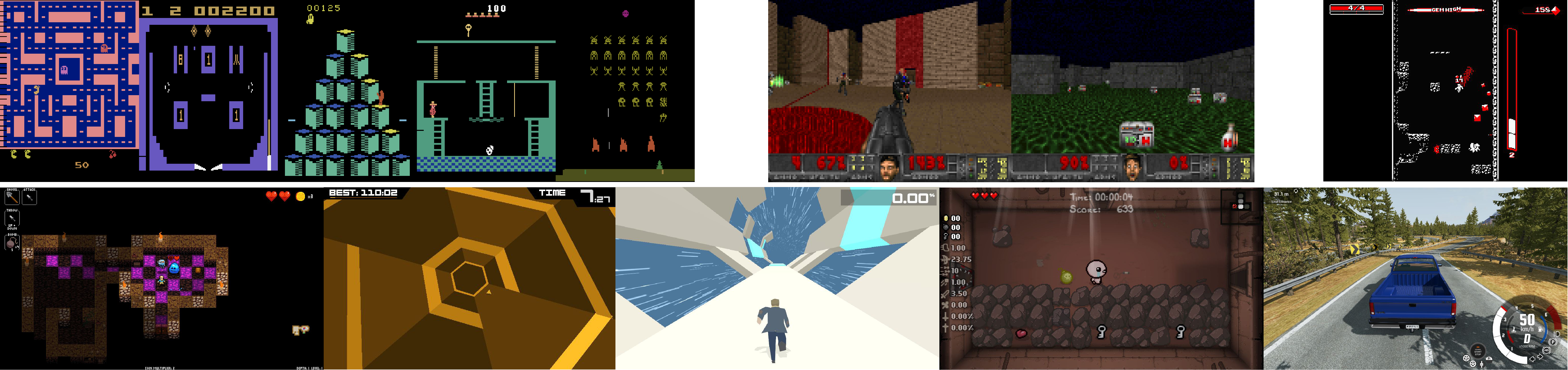}
    \caption{Games tested with behavioural cloning. Images represent what the BC agent would see. From left to right: Ms. Pac-Man, Video Pinball, Q*bert, Montezuma's Revenge, Space Invaders, Deathmatch (Doom), HGS (Doom), Downwell, Crypt of The NecroDancer, Super Hexagon, Boson X, Binding of Isaac: Rebirth and BeamNG.drive.}
    \label{fig:games}
    \end{figure*}

    \textit{Reinforcement learning} (RL) \cite{sutton2018reinforcement} has been successfully applied to create super-human players in multiple video games, including classic Atari 2600 games \cite{dqn}, as well as more modern shooters \cite{ctf}, MOBAs \cite{five, ye2019mastering} and real-time strategy games \cite{vinyals2019grandmaster}. Even more so, all before-mentioned accomplishments use \say{end-to-end} systems, where input features are not pre-processed by crafting specific features, and instead rely on raw information like image pixels. However, RL is not without its limitations: they require an environment where to play the game. Whether this is achieved by modifying an existing game (like Starcraft II \cite{sc2}) or by using their engines to create environments from ground-up (like Unity ML-Agents \cite{unityml}), it still requires considerable engineering. Even worse, after the environment is created, training the agents may take thousands of years of in-game time \cite{five, vinyals2019grandmaster}. 
    
    An alternative approach is \textit{imitation learning}, in which agents learn to replicate demonstrators' actions. \textit{Behavioural cloning} (BC) \cite{pomerleau1989alvinn} is the simplest form of this: given an observation and an associated action from a demonstrator, predict this action based on observation (\textit{i.e} a classification task). This has been used to kick-start RL agents \cite{vinyals2019grandmaster, hester2018deep}, but also applied alone in \textit{e.g.} autonomous driving \cite{pomerleau1989alvinn, bojarski2016end, codevilla2018end}, and Vinyals \textit{et al.} \cite{vinyals2019grandmaster} show that Starcraft II can be played at proficient human-level with behavioural cloning alone. This begs the question: \textbf{How well can behavioural cloning play video games, in general?} Can we reach the level of a human player? How much data do we need? Do we need data from multiple players?

    If we can create performant, end-to-end agents with BC and human gameplay alone, it would skip many hurdles experienced with RL: we do not need to create an environment for agents to play in, nor do we need to spend large amounts of compute resources for training. We only need the video game, a tool to record the gameplay, and players for the game. If BC can manage with just an hour or two of gameplay demonstration, a single person could record the demonstration data. If the recording tool captures the same output and input a human player would have (\textit{i.e.} image of the screen and keyboard/mouse, \textit{end-to-end}), this would require no game-specific coding and could be applied to any game. Even if BC does not reach human-level performance, it could still be used as a starting point for other learning methods, or as a support for diversifying the agent's behaviour \cite{vinyals2019grandmaster}.
    
    Video games have been in active use as benchmarks in research using BC \cite{hester2018deep, guss2019minerldata, zhang2019atari, kurin2017atari}, and as milestones to beat in AI research \cite{dqn, ye2019mastering, vinyals2019grandmaster}. The other way around, \say{BC for video games}, has seen works like human-like bots in first-person shooter (FPS) games using hand-crafted features and imitation learning \cite{pelling2019two, zanetti2004machine, gorman2007imitative}, end-to-end FPS bots with RL and BC \cite{harmer2018imitation}. Our setting and motivation resemble the motivation of \cite{chen2017game}, where authors employ end-to-end imitation learning to play two Nintendo 64 games successfully. However, these studies have been limited to only a few games a time, making it hard to tell how well BC performs in general at playing video games. Apart from \cite{kurin2017atari}, related work does not study how data should be chosen for behavioural cloning. In addition, Zhang \textit{et al.} \cite{zhang2019atari} bring up an important point on how human delay can adversarially affect the quality of the dataset but did not include experimental results on this.

    In this work, we aim to answer these three questions and to assess the general applicability of end-to-end behavioural cloning for video game playing. We use data from human demonstrators to train a deep network to predict their actions, given the same observations human players saw (the screen image). We run empirical experiments to study how well BC agents play Atari 2600 games, Doom (1993) and various modern video games. Along with the raw performance, we study the effect of quality and quantity of the training data, and the effect of delay of human reflexes on the data quality. Along the results, we present  \textit{ViControl} (\say{Visual Control}), a multi-platform tool to record and play an arbitrary game, which we use to do behavioural cloning on the modern games. ViControl is available at \url{https://github.com/joonaspu/ViControl}.

\section{End-to-end behavioural cloning for video games}
    \subsection{Behavioural cloning}
        \label{sec:bc}
        We wish to train computer to play a game, based on given demonstrations of humans playing it. We model the environment as a truncated version of Markov Decision Processes (MDPs) \cite{sutton2018reinforcement}, where playing the game consists of observations $s \in \mathcal S$ and associated actions $a \in \mathcal A$. We do not include a notion of time, reward signal nor terminal/initial states. The task of behavioural cloning is simple: given a dataset of human gameplay $\mathcal D$ containing tuples $(s, a)$, learn the conditional distribution $p(a | s)$, \textit{i.e.} probability of human players picking action $a$ in state $s$. After learning this distribution, we can use it to play the game by sampling an action $a \sim p(a | s)$ for a given state $s$ (\textit{agent}). An immediate limitation here is the lack of temporal modelling, or \say{memory}, which could limit the agent's abilities. It has been shown that including past information with behavioural cloning can be detrimental to performance \cite{de2019causal}, but on the other hand there exists work that successfully do BC with recurrent neural networks \cite{scheller2020sample}. We opt not to use recurrent networks for the model and training simplicity, and as most of the games used in this work do not require memory to master.
        
        We take an end-to-end approach, where states are pixels of an RGB image $s \in \mathbb R ^ {H \times W \times 3}, \; H, W \in \mathbb N$, and actions $\mathbf a$ are a vector of one or more discrete variables $a_i \in {0, 1, \ldots d_i}$, where $i \in \mathbb N$ represents the number of discrete variables, and $d_i$ tells the number of options per discrete variable. In Atari environments \cite{bellemare2013arcade}, action contains one discrete variable with $18$ options, including all possible choices human player could make (multi-class classification task). With a PC game using a keyboard with, say, four buttons available, the actions consist of four discrete variables, all with two options: \texttt{down} or \texttt{up} (multi-label classification task).
        
        To model the conditional distribution $p(a|s)$, we use deep neural networks. They support the different actions we could have and are known to excel in image classification tasks \cite{lecun2015deep}. We treat action discrete variables $i$ independent from each other, and the network is trained to minimize cross-entropy between predictions and labels in the dataset. 
        

    \subsection{Challenges of general end-to-end control of video games}
        \label{sec:challenges}
        Compared to Atari 2600 games and Doom (1993), modern video games (published after 2010) can take advantage of more computing power and tend to be more complex when it comes to visual aesthetics and dynamics. We also do not assume to have control over game program's flow, so the game will run at a fixed rate, as humans would experience it. All-together, these raise some specific challenges for generalized end-to-end control, where we wish to avoid per-game engineering.

        \paragraph{High resolution}
            Modern games commonly run at \say{high definition} resolutions, with most common monitor resolution for players being $1920 \times 1080$. However, RL and BC agents resize images to small resolutions due to computational efficiency, usually capped around $200$ to $300$ pixels per axis \cite{ctf, vizdoom_competitions}, and commonly lower \cite{dqn, hester2018deep}. If we take a modern game with a resolution of at least $1280 \times 720$, and downscale it to these resolutions, we lose a great deal of detail: any smaller user-interface (UI) elements, like text, may get blurred out, and already-small objects on the screen may disappear completely. On top of this, different interpolation methods used for resizing images have been reported to affect the training results \cite{bukatyusing, chen2017game}. We leave approaches for solving this to future work, and simply resize the images.

        \paragraph{Complex action space}
            The natural action space of a computer game, a keyboard and a mouse, contains over a hundred keys to press in total, as well as the movement of the mouse. Such large action spaces have shown to be an issue to RL agents \cite{dulac2015deep, zahavy2018learn}, and many of these buttons do nothing in games (when was the last time you have used \verb|Insert| in a video game?). Even when we modify the action space to only include buttons that are used by the game, we can end up with a large, parametrized action space with its own difficulties \cite{delalleau2019discrete}, like in Starcraft II \cite{sc2}. We pre-define a minimal set of actions required to play games in this work.
        
        \paragraph{Asynchronous execution}
            As the game environment runs asynchronously from the agent's decisions, the agent must execute an action quickly after observing an image, otherwise its decisions will lag behind. This \say{control delay} is known to reduce performance of RL methods \cite{firoiu2018human, schuitema2010control}. In addition, if we gather BC data from human players, the recorded actions are subject to delays from human-reflexes. If something surprising happens on the screen, average humans react to this with a split-second delay. This action was supposed to be associated with the surprising event, but instead it will be recorded few frames later, associated with possibly a wrong observation. Other way around, human players could plan their action before the observation and execute it pre-maturely. These both lead to \textit{state-action mismatch} \cite{zhang2019atari}, effect of which we study in the experiments.
            
        
        \paragraph{Confounding information}
            Behavioural cloning is prone to \textit{causal confusion} \cite{de2019causal} or \say{superstition} \cite{bontrager2019superstition}, where providing more information may be detrimental to BC agent's performance. With more information (\textit{e.g.} history, past frames/actions), the model has a larger chance to find misleading correlations between observations and actions. For example, firing a plasma-weapon in Doom. This creates a blue, long-lasting muzzle-flash on the weapon. Since many frames with \texttt{ATTACK} pressed down include this blue flash, the model learns to focus on this flash to predict if we should fire the weapon. However, the flash is not the \textit{cause} of firing the weapon, it is the \textit{effect}. Similarly, games have numerous UI elements with various information, which could lead to similar confusion. In this work we do not provide historical information to the agent, limiting its capabilities in exchange for less chance of destructive causal confusion.

\section{Research questions and experimental setup}
    \begin{figure}
        \centering
        \includegraphics[width=0.95\columnwidth]{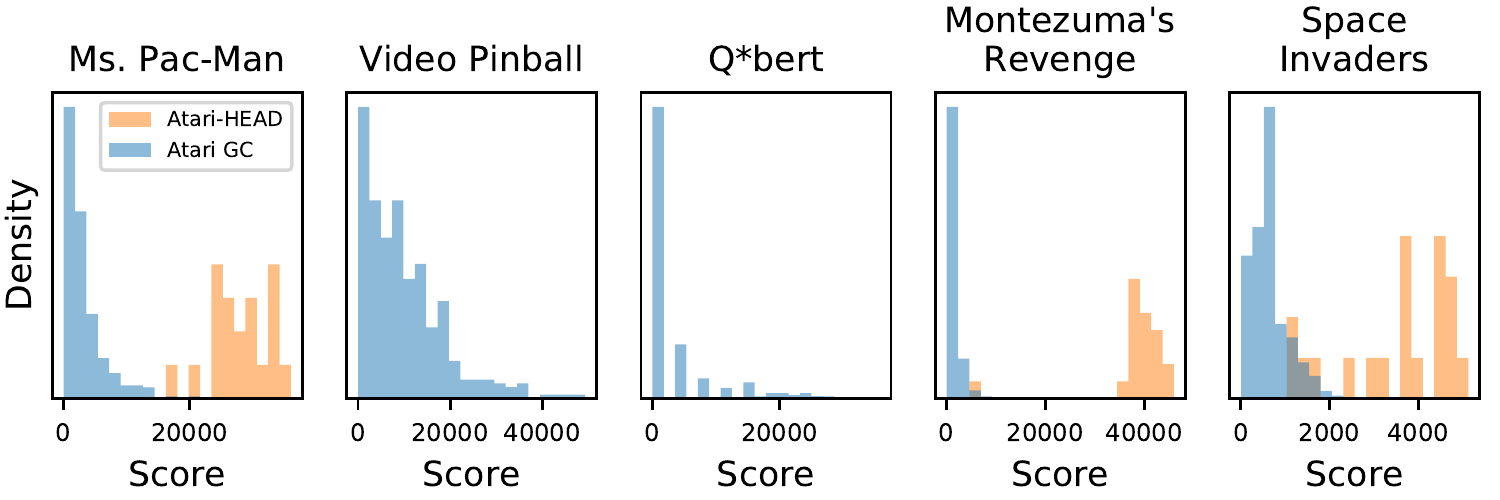}
        \caption{Comparison of the score distribution in the Atari Grand Challenge (Atari-GC) dataset and in the Atari-HEAD dataset. Atari-HEAD does not have data for \textit{Video Pinball} or \textit{Q*bert}}
        \label{fig:atari-histogram}
    \end{figure}
    
    Along with the main evaluation of BC in different games, we study two important aspects of the training setup, brought up by related work: \say{how does the quantity and quality of the data affect the results?} \cite{kurin2017atari}, and \say{how the state-action mismatch from human reflexes affects the results?} \cite{zhang2019atari}. The former sheds light on if we should gather data from only few experts, or should we use data from many different players. A similar comparison of different sized datasets was done in \cite{kurin2017atari}. The latter was brought up by authors of \cite{zhang2019atari}, but without experimental results. 
    
    To study the state-action mismatch, we run experiments with modified versions of the Atari and ViZDoom datasets, where an \textit{action delay} $d$ is added between the state and action. In the modified datasets, the state $s_i$ at frame $i$ is matched with an action $a_{i + d}$. Both positive and negative values are used for the action delay $d$.
    
    We will use Atari \cite{bellemare2013arcade} and Doom \cite{vizdoom} environments to answer these questions, as they can be used synchronously and therefore allow fast evaluation of trained models. We will then include six modern video games to assess how well BC works under the challenges presented in Section \ref{sec:challenges}. Images of all of the games are shown in Figure \ref{fig:games}. Code used to run these experiments is available at \url{https://github.com/joonaspu/video-game-behavioural-cloning}.
    
    \subsection{Evaluation}
        For the Atari and Doom experiments, each training run is evaluated by taking models from the three last epochs of training, evaluating their performance and averaging over. The training is then repeated three times with random seeds, and the result shown is an average over these three runs. We do this to capture the variance between different training steps, illustrated in Figure \ref{fig:atari-training}. The same is done when evaluating with the modern games, except we only evaluate the final model instead of the last three epochs. 
        
        The evaluation results are reported as percentage of human score \cite{dqn}, where 0\% is a baseline score set by an agent that picks a random action on each frame, and 100\% is the mean score of the human players in the dataset used for training. In Atari experiments, we use the mean score of the episodes with a score above the 95th percentile in the Atari Grand Challenge dataset \cite{kurin2017atari} for average human score.
    
    \subsection{Behavioural cloning model}
        The neural network model is based on the convolutional neural network used in the original deep Q-learning work \cite{dqn} and in related BC experiments \cite{zhang2019atari}, consisting of three convolutional layers, followed by a single fully connected layer of $512$ units and a layer that maps these to probabilities for each action. All layers use ReLU (rectified linear unit) activations. While small by modern standards, this architecture is the \textit{de facto} architecture used in RL experiments \cite{hester2018deep, stable-baselines}. Residual networks \cite{he2016deep} have also been used for improved performance \cite{espeholt2018impala, bukatyusing}, but are slower to run. We opt for the faster, simpler network to keep up with the fast pace of actions required for experiments with asynchronous games, described in Section \ref{sec:modern-games}. All code is implemented in PyTorch.
        
\begin{figure}
        \centering
        \includegraphics[width=0.95\columnwidth]{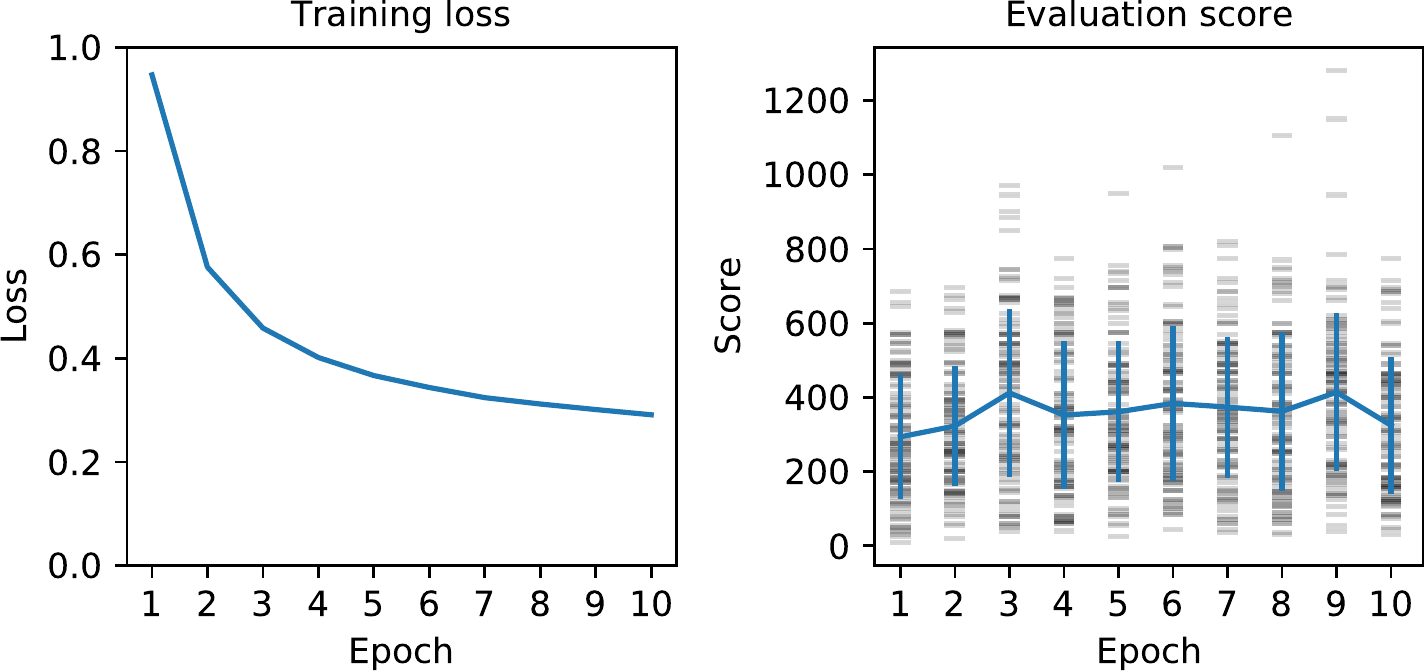}
        \caption{An example of the training loss (left) and evaluation score curves (right). Black lines are scores of individual evaluation episodes. Plots are from training on Space Invaders on the Atari-HEAD dataset and are similar across different games.}
        \label{fig:atari-training}
        \end{figure}

        In all of the experiments, the network is trained using the Adam optimizer \cite{adam} to minimize cross-entropy, with a learning rate of $0.001$ and L2-normalization weight $10^{-5}$. In all experiments, we train until training loss does not improve. We did not find shorter or longer training to be helpful, and agent's performance did not increase significantly after first $10-50\%$ of the training regimen (see Figure \ref{fig:atari-training}). Interestingly, training loss continues to reduce while evaluation performance does not change. This is expected to a degree, as the training loss (per-sample prediction error) does not reflect the agent's performance \cite{ross2011reduction}. During evaluation, we sample the final actions according to the probabilities provided by the network. We found this to work better than deterministically selecting the action with the highest probability.
        
    \subsection{Atari games}
        \begin{table}
            \centering
            \caption{Statistics for the Atari Grand Challenge and Atari-HEAD datasets}
            \label{fig:atari-dataset-stats}
            \begin{tabular}{l|cc}
            Environment and dataset                     & Episodes  & Total samples \\ \hline
            \textbf{Ms. Pac-Man}                        &           &               \\
            \hspace{3mm}Atari Grand Challenge, All data & $667$     & $2829068$     \\
            \hspace{3mm}Atari Grand Challenge, Top 50\% & $335$     & $2066077$     \\
            \hspace{3mm}Atari Grand Challenge, Top 5\%  & $34$      & $362056$      \\
            \hspace{3mm}Atari-HEAD                      & $20$      & $353428$      \\
            
            \textbf{Video Pinball}                      &           &               \\
            \hspace{3mm}Atari Grand Challenge, All data & $380$     & $2352787$     \\
            \hspace{3mm}Atari Grand Challenge, Top 50\% & $190$     & $1688256$     \\
            \hspace{3mm}Atari Grand Challenge, Top 5\%  & $19$      & $224150$      \\
            
            \textbf{Q*bert}                             &           &               \\
            \hspace{3mm}Atari Grand Challenge, All data & $1136$    & $3329088$     \\
            \hspace{3mm}Atari Grand Challenge, Top 50\% & $576$     & $2419198$     \\
            \hspace{3mm}Atari Grand Challenge, Top 5\%  & $57$      & $614193$      \\
            
            \textbf{Montezuma's Revenge}                &           &               \\
            \hspace{3mm}Atari Grand Challenge, All data & $1196$    & $4623879$     \\
            \hspace{3mm}Atari Grand Challenge, Top 50\% & $931$     & $3991548$     \\
            \hspace{3mm}Atari Grand Challenge, Top 5\%  & $92$      & $646985$      \\
            \hspace{3mm}Atari-HEAD                      & $20$      & $335276$      \\
            
            \textbf{Space Invaders}                     &           &               \\
            \hspace{3mm}Atari Grand Challenge, All data & $905$     & $4005345$     \\
            \hspace{3mm}Atari Grand Challenge, Top 50\% & $483$     & $2765214$     \\
            \hspace{3mm}Atari Grand Challenge, Top 5\%  & $46$      & $422372$      \\
            \hspace{3mm}Atari-HEAD                      & $20$      & $332483$
            
            \end{tabular}
        \end{table}
    
        For the Atari experiments, we used two existing datasets: the Atari Grand Challenge dataset (\textit{Atari GC}) \cite{kurin2017atari} and the Atari-HEAD dataset \cite{zhang2019atari}. The Atari Grand Challenge dataset includes five games, which were all used for our experiments. The Atari-HEAD dataset includes a set of $20$ Atari games, out of which we used the three games that are also in the Atari Grand Challenge dataset. Atari-HEAD includes episodes with higher score. A comparison of the distribution of final scores in these two datasets can be seen in Figure \ref{fig:atari-histogram}.
        
        
        To study effect of the amount and quality of the data on behavioural cloning, we include experiments similar to ones in \cite{kurin2017atari}, where we repeat behavioural cloning only using episodes with scores above the 95th percentile and 50th percentile (\say{top 5\%} and \say{top 50\%}). It should be noted that we use only BC, while \cite{kurin2017atari} used deep Q-learning from demonstrations \cite{hester2018deep}. The amount of data for all these setups are shown in Table \ref{fig:atari-dataset-stats}. 
        
        
        In both datasets, the input frames are $160 \times 210$ RGB images that are resized to $84 \times 84$ when training the models. To eliminate flickering of certain in-game elements, each frame is merged with its preceding frame by setting each pixel to have the lighter value from these two frames (maximum). The frame rate in both datasets is $60$ frames per second. Models were trained for $10$ epochs, except for the full Atari GC dataset and its top 50\% subset, which were trained for $5$ epochs.
    
        The models are evaluated with the OpenAI Gym Atari environments with $100$ episodes, with default environments (\say{v4} versions). Evaluation runs until the game ends or the $40000$th frame is reached.
        
        
    \subsection{Doom}
        For the Doom experiments, we use two scenarios provided by the ViZDoom \cite{vizdoom}: \textit{Health-Gathering-Supreme} (HGS) and \textit{Deathmatch}. In both scenarios the input observation is an RGB image of size $80 \times 60$, and the network predicts which of the allowed buttons are pressed down. Human gameplay is recorded every other ViZDoom tick ($17.5$ frames per second), and the trained model takes actions at the same rate. We collect data from three players, and train models for $30$ epochs. Evaluation is done the same way as with the Atari experiments, except with $200$ games per epoch.

        In the HGS scenario, the player constantly takes damage, and must navigate around a small maze to collect med kits to survive longer. The longer the player survives, the higher the score. Allowed buttons are \verb|TURN_LEFT|, \verb|TURN_RIGHT| and \verb|MOVE_FORWARD|. The game ends when the player dies or a timeout (one minute) is reached. We record $20$ full games per person, totaling around one hour of gameplay and $62615$ samples.
        
        In the Deathmatch scenario, the player is pitted against a room of randomly spawning enemies, with a generous number of pickups and weapons on the sides of the levels. Allowed buttons are \verb|ATTACK|, \verb|SPEED| (hold down for faster movement), \verb|TURN_LEFT|, \verb|TURN_RIGHT|, \verb|MOVE_FORWARD| and \verb|MOVE_BACKWARD|. The game ends when the player dies, or a timeout of two minutes is reached. We record $10$ games per person, with total of $46243$ samples, corresponding to $45$ minutes of gameplay.
        
    \subsection{Modern video games}
        \label{sec:modern-games}
        As for the experiments with modern video games (released after 2010), we selected games that are familiar to players who provide the data, and which do not require a mouse to play. The selected games are described in Appendix \ref{appendix:games}, with example images in Figure \ref{fig:games}. We use a specifically built tool, ViControl, to capture the screen image, the corresponding keyboard/mouse input, and to later emulate these buttons to allow the agent to play the game. During recording, ViControl behaves like any game recording/streaming software, except it also tracks keypresses. We collect data from two players, with $30$ minutes of gameplay from both, totaling $\approx 72 000$ frames of demonstration per game. Models were trained for 30 epochs. The only pre-processing we apply is resizing the image. Evaluation is done by letting the trained model play the game until the game ends, the episode lasts too long, or when some other game-specific criteria is met. The final score is an average over ten such games.


\section{Results and Discussion}
    
    \begin{table*}
    \centering
    \caption{Results with behavioural cloning. Our score is the highest average score over different dataset sizes and action-delays used. Variances are not included as they differ from work to work (we report variance over multiple training runs, Zhang et al. 2019 reports variance over multiple episodes).
    }
    \label{fig:results-comparison}
    \begin{tabular}{l|ccccc}
    Game                       & Random agent & Human Average & Behavioural cloning (our)        & Hester et al. 2018 & Zhang et al. 2019 \\
    \hline
    Ms. Pac-Man                & $173.3$      & $12902.5$     & $811.7$ (GC, All, +2 delay)           & $692.4$            & $1167.5$          \\
    Video Pinball              & $22622.4$    & $34880.1$     & $21715.0$ (GC, top 5\%)          & $10655.5$          & N/A               \\
    Q*bert                     & $162.9$      & $23464.0$     & $9691.6$ (GC, top 5\%, +5 delay) & $5133.8$           & N/A               \\
    Montezuma's Revenge        & $0.2$        & $4740.2$      & $1812.1$ (GC, top 5\%, +5 delay) & $576.3$            & $970.2$           \\
    Space Invaders             & $158.8$      & $1775.9$      & $564.9$ (HEAD, +2 delay)         & N/A                & $247.1$           \\ \\
    Health Gathering (ViZDoom) & $3.1$        & $20.9$        & $9.4$ (+2 delay)                 &                    &                   \\
    Deathmatch (ViZDoom)       & $2.5$        & $93.1$        & $13.1$ (+2 delay)                &                    &                   \\ \\
    Downwell                   & $92$         & $1054.8$      & $81.2$                           &                    &                   \\
    Crypt of the NecroDancer   & $0$          & $440.4$       & $4.0$                            &                    &                   \\
    Super Hexagon              & $3.3$        & $112.5$       & $4.6$                            &                    &                   \\
    Boson X                    & $0$          & $170.7$       & $2.4$                            &                    &                   \\
    Binding of Isaac: Rebirth  & $287.6$      & $2045.8$      & $463.4$                          &                    &                   \\
    BeamNG.drive               & $27.8$       & $3525$        & $477.1$                          &                    &                  
    \end{tabular}
    \end{table*}
    
    \subsection{General behavioural cloning performance}
        \begin{figure}
        \centering
        \includegraphics[width=0.95\columnwidth]{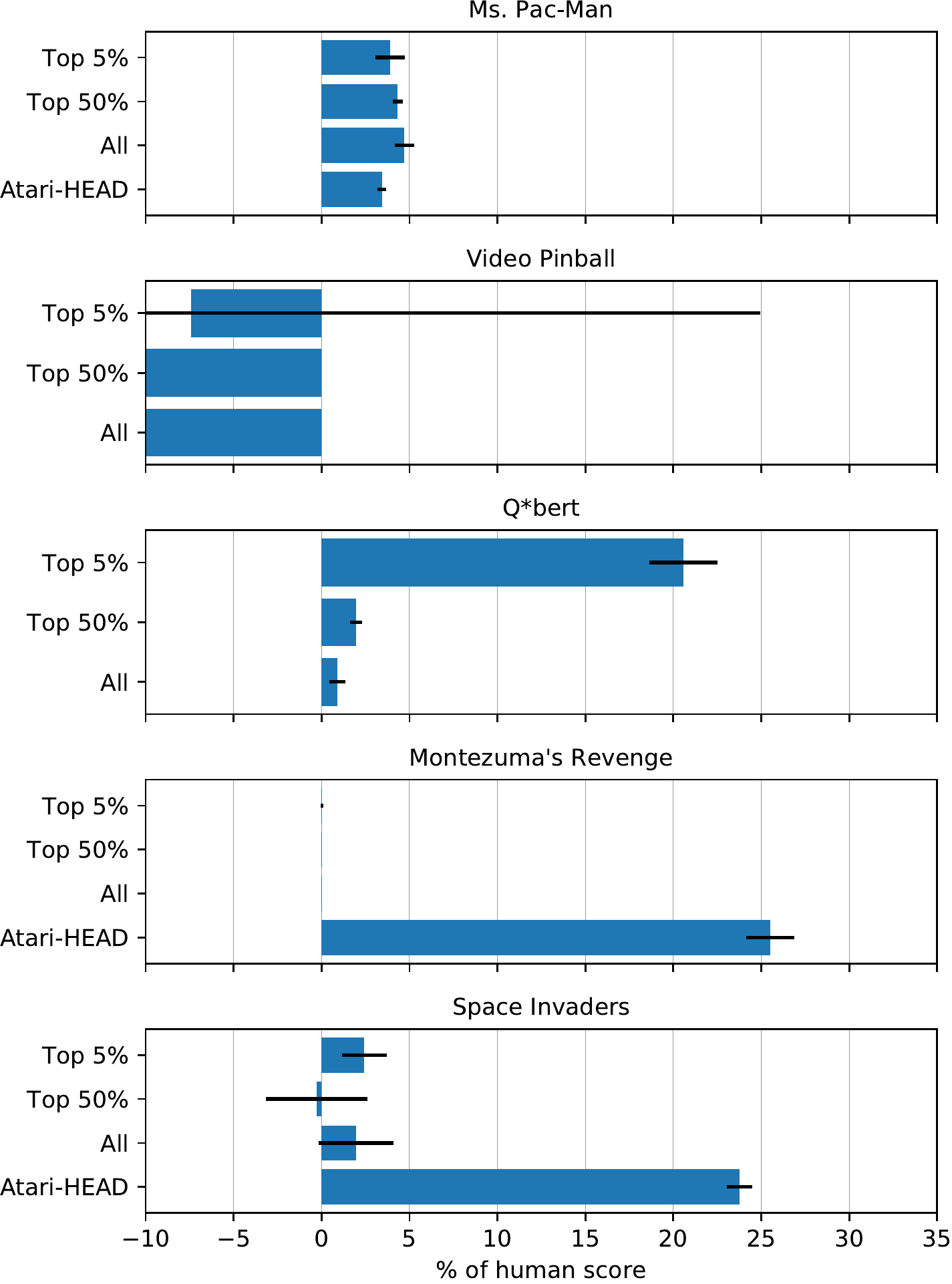}
        \caption{Human-normalized scores of behavioural cloning on the three different subsets of Atari Grand Challenge dataset and for the Atari-HEAD dataset.}
        \label{fig:atari-plot}
        \end{figure}
    
        Figure \ref{fig:atari-plot} shows the results for the model trained with both Atari datasets. Ms. Pac-Man results show a fairly poor performance of under 5\% of human score. Video Pinball fails to achieve the baseline score set by a random agent. Q*bert, Montezuma's Revenge and Space Invaders, however, reach a score of over 20\% of human score.
        
\begin{figure}
        \centering
        \includegraphics[width=0.95\columnwidth]{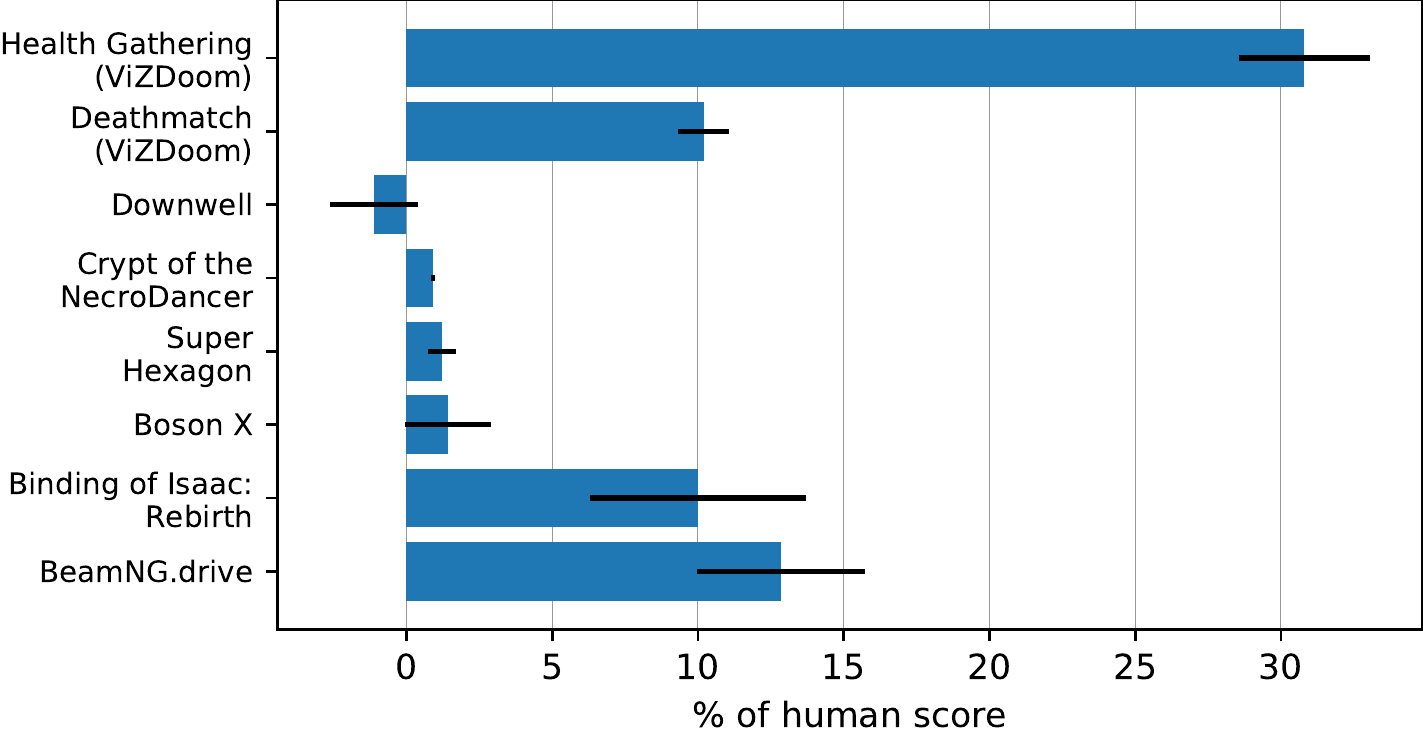}
        \caption{Human-normalized scores of behavioural cloning on the two ViZDoom scenarios and the six modern video games.}
        \label{fig:all-games-plot}
        \end{figure}

        The results in Figure \ref{fig:all-games-plot} show the performance of the two ViZDoom scenarios as well as the modern video games. Out of these, ViZDoom health gathering is the only one to achieve a human normalized score of more than 30\%, while others remain under 15\%. Out of the modern video games, Binding of Isaac: Rebirth and BeamNG.drive are the only games that get a score significantly above the baseline set by a random agent.
        
        Despite the low scores in most tested games, watching the agents' gameplay shows that the models still learn some of the basic dynamics of the games. See video available at \url{https://youtu.be/2SMLpnUEIPw}. In Super Hexagon, the agent moves in the correct direction, but often overshoots or undershoots the correct position. In Binding of Isaac: Rebirth, the agent moves through doors and shoots towards enemies and in BeamNG.drive, the agent accelerates and steers in the correct direction, but still hits the walls and damages the car often. In Boson X, agent learns to jump at the right moments, but often jumps too short to reach the other platforms. In Crypt of the NecroDancer, the agent learns to hit nearby enemies and move in the tunnels, but often throws away their weapon or kills themselves with a bomb.
        
        Comparing our results with earlier BC experiments done by Hester \textit{et al.} \cite{hester2018deep} and Zhang \textit{et al.} \cite{zhang2019atari} (Table \ref{fig:results-comparison}) we reached higher scores in all tested Atari games except for Ms. Pac-Man, by adjusting for human action-delay and only using higher quality data. The results in Kurin \textit{et al.} \cite{kurin2017atari} are not directly comparable, since they did not use a pure BC method.
        
    \subsection{Data quality versus quantity}
        
        \begin{figure*}
            \centering
            \includegraphics[width=0.95\textwidth]{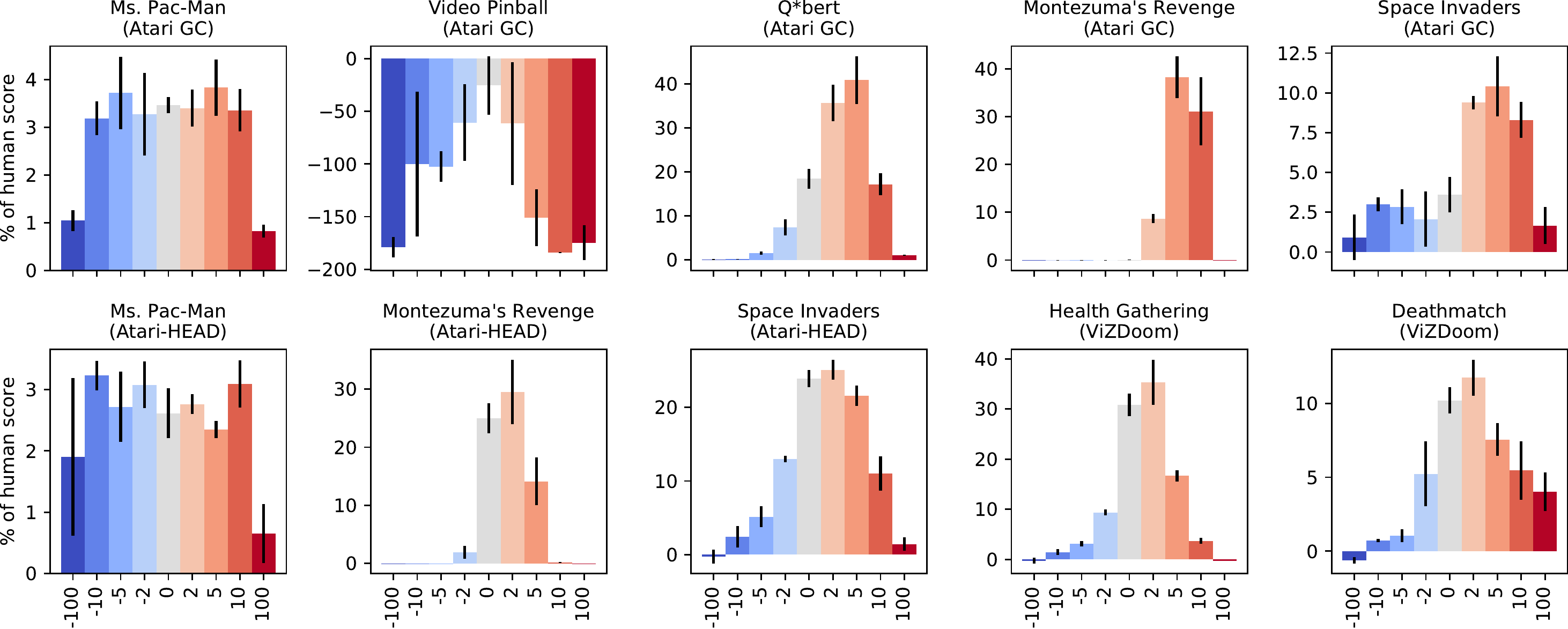}
            \caption{Results with action delay of Atari Grand Challenge (top 5\%), Atari-HEAD and ViZDoom datasets. X-axis represents the action-delay used while training the model, with positive meaning the action lags behind. \textit{E.g.} delay of five means we move all actions five steps back in time, and associate with corresponding observation.}
            \label{fig:delay-plot}
        \end{figure*}
    
        Looking more closely at the Atari results in Figure \ref{fig:atari-plot} we can see that Q*bert and Space Invaders benefit significantly from having smaller but higher quality training datasets. Q*bert score increases from just barely above the random agent's performance to over 20\% of human score when using the top 5\% of episodes. Space Invaders gets a similar increase when moving from the Atari Grand Challenge dataset to the Atari-HEAD dataset. Differences in Ms. Pac-Man are not significant, given the small change and relatively large variance.
        
        To further study the effect that the quantity of data has on the results, we ran experiments with datasets that only contained the top 1, 2 and 3 episodes of the Atari Grand Challenge dataset. In many games the results were still comparable to results shown here, considering the very small amount of data. For example, Ms. Pac-Man got a score of $515$ with just the best two episodes ($28330$ samples) of the dataset. Training with the entire dataset ($2829068$ samples) resulted in a score of $774$. The score with the top two episodes of Space Invaders ($20112$ samples) was $193$, while a model trained with the full dataset ($4005345$ samples) got a slightly lower score of $190$ points. Q*bert score, however, dropped sharply when smaller datasets than the top 5\% were used. These results suggest that even a very small amount of high-quality data can result in a comparatively well performing agent.
        
        For Doom experiments, we trained models with each player's data, as well as with all players' data combined. On HGS (Health Gathering Supreme), an agent trained with the data collected from one of the players achieved a slightly higher score than the agent trained with all players' combined data. With the Deathmatch scenario, however, the agent trained with the combined data reached a higher score than any of the agents trained with individual players' data. We believe this is because of the complexity of the two scenarios: deathmatch has a wide variety of different enemies and available weapons and items, so having more data is beneficial. HGS is a more straightforward scenario. Interestingly, despite all three players had highest score possible in HGS in all of the recorded data, the performance of trained agents varied between $4$ and $11$ average score, depending on which player's data agent was trained on. We believe this is because of the differences in how different participants played the game.

    \subsection{Action delay}
    
        The first row of Figure \ref{fig:delay-plot} shows the action delay results for the Atari Grand Challenge dataset. Q*bert, Montezuma's Revenge and Space Invaders see a significant increase in evaluation scores with positive action delay values, with the largest increase seen when using a delay of five frames. Action delay does not have a large effect with Ms. Pac-Man, apart from a large drop in final score caused by delay values of $-100$ and $100$. Video Pinball achieves the best performance with zero action delay, although the score is still well below the 0\% mark set by the random agent. Results for Atari-HEAD dataset show smaller yet consistent improvements with two frame delay in Montezuma's Revenge and Space Invaders. Same applies to both ViZDoom scenarios, where delay of two improved the performance slightly over zero delay.

        With the Atari games' frame rate of $60$, the delay of five frames (Atari-GC) corresponds to about $83$ milliseconds of delay, and two frames (Atari-HEAD) is about $33$ milliseconds. Our ViZDoom datasets are collected at $17.5$ frames per second, and delay of two frames corresponds to $114$ milliseconds. The differences between two Atari datasets reflect how Atari-HEAD was collected in a synchronous manner (game waited for human to execute an action), with delay from observation to action being lower albeit not zero.
    
\section{Conclusion}
    We benchmarked end-to-end behavioural cloning in various video games and studied the effect of quality of expert data and the delay from human reaction time. Our results show that behavioural cloning agents can learn basic mechanics/rules of the games (\textit{e.g.} coherent movement) with a small amount of data (one hour of human gameplay), but generally only achieve fraction of the performance of human players, and sometimes even worse than a random agent. We demonstrate how the quantity of the data matters less when only a limited amount of data is available, and how adjusting for the human delay from observations to actions (reflexes) improves the performance.
    
    Based on these results, we recommend using high-quality data, rather than just large quantities of any data, for behavioural cloning. If data is gathered from human demonstrators, we also recommend offsetting the recorded action by assigning them to observations $100\text{ms}$ earlier. This is to counteract the state-action mismatch introduced by the delay from observations to actions.
    
    As a future work, we would like to solve  issues that still remain, \textit{e.g.} using \say{super-resolution} networks to handle high-definition images instead of resizing them, using recurrent networks and trying to avoid causal confusion \cite{de2019causal}. A simple question remaining is also how far we can get with BC with a large amount of data, like in Starcraft II \cite{vinyals2019grandmaster}. Going beyond BC, methods like generative adversarial imitation learning (GAIL) \cite{ho2016generative} and batch reinforcement learning \cite{fujimoto2019benchmarking} require simulations or reward signals but show improvements over behavioural cloning. All things considered, including successful applications of reinforcement learning and the recent improvements in imitation learning, we remain hopeful for human-level agents in video games, despite the less-than-ideal results presented here.

\bibliographystyle{ieeetr}
\bibliography{main}

\appendices
\section{Game descriptions}
\label{appendix:games}
    \paragraph{Downwell}
        A roguelike, vertically scrolling platformer published by Devolver Digital in 2015, with simple dynamics and graphics. Human players were instructed not to use shops, as buying items affects the final score.
        
        \begin{itemize}
            \item Resolution: $760 \times 568$ ($95 \times 70$)
            \item Allowed buttons: \texttt{jump/shoot}, \texttt{left} and \texttt{right}.
            \item Game start: Start of the game (when player selects \say{restart}).
            \item Game end: Player death or $5$ minute timeout.
            \item Score: Number of gems upon end of the game.
        \end{itemize}
    
    \paragraph{Crypt of The NecroDancer (CoTN)}
        A roguelike, rhythm-based dungeon exploration game published by Brace Yourself Games in 2015. Normally, players and NPCs move only at the beats of the music, but we remove this mechanic by using an easier character (\say{Bard}), to focus on the dungeon exploration aspect. Human players were instructed not to use shops, as buying items affects the final score.
        
        \begin{itemize}
            \item Resolution: $1280 \times 720$ ($160 \times 90$).
            \item Allowed buttons: \texttt{left}, \texttt{right}, \texttt{up} and \texttt{down}.
            \item Game start: Start of the \say{all floors run}.
            \item Game end: Death, reaching Zone 2 or $10$ minute timeout.
            \item Score: Number of coins in the end.
        \end{itemize}
    
    \paragraph{Super Hexagon}
        A 2D \say{twitch} video game, where player has to simply avoid incoming obstacles, published by Terry Cavanagh in 2012.
        
        \begin{itemize}
            \item Resolution: $1280 \times 720$ ($160 \times 90$).
            \item Allowed buttons: \texttt{left} and \texttt{right}.
            \item Game start: Start of the first level (\say{Hexagon}, normal mode).
            \item Game end: Death.
            \item Score: Time survived in seconds.
        \end{itemize}
    
    \paragraph{Boson X}
        A 3D twitch game by Ian MacLarty (2014), where player has to jump over holes and obstacles in speeding-up platform.
        
        \begin{itemize}
            \item Resolution: $1280 \times 720$ ($160 \times 90$).
            \item Allowed buttons: \texttt{left} and \texttt{right}.
            \item Game start: Start of the first level (\say{Geon}).
            \item Game end: Death.
            \item Score: In-game score.
        \end{itemize}
    
    \paragraph{Binding of Isaac: Rebirth (BoI)}
        A roguelike, top-down shooter published by Nicalis Inc. in 2014 (a remake of \say{Binding of Isaac}), where player progresses in rooms by killing all the enemies and collecting items to power themselves up. In-game score ticks down as time progresses, but we use it to include activity of the player.
        
        \begin{itemize}
            \item Resolution: $1280 \times 720$ ($160 \times 90$).
            \item Allowed buttons: \texttt{left}, \texttt{right}, \texttt{up}, \texttt{down}, \texttt{shoot left}, \texttt{shoot right}, \texttt{shoot up}, \texttt{shoot down} and \texttt{place bomb}.
            \item Game start: Start of the game with \say{Isaac} character with default settings.
            \item Game end: Death, beating the second boss or $10$ minute timeout.
            \item Score: In-game score.
        \end{itemize}
    
    \paragraph{BeamNG.drive}
        A driving game with accurate models of car mechanics, published by BeamNG in 2015.
        
        \begin{itemize}
            \item Resolution: $1280 \times 768$ ($165 \times 96$).
            \item Allowed buttons: \texttt{accelerate}, \texttt{brake}, \texttt{left}, \texttt{right}.
            \item Game start: The \say{Handling Circuit} spawn point on the \say{Automation Test Track} map with the \say{Gavril D-Series D15 V8 4WD (A)} vehicle.
            \item Game end: Two full laps completed, or agent does not move for $10$ seconds (\textit{e.g.} stuck, car immobilized).
            \item Score: Meters driven until a collision or the end of the second lap (as reported by the in-game \say{Trip Computer}).
        \end{itemize}
\vspace{12pt}

\end{document}